# scientific reports

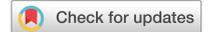

### OPEN

# An adaptive shortest-solution guided decimation approach to sparse high-dimensional linear regression

Xue Yu[1], Yifan Sun[1]✉ & Hai-Jun Zhou[2,3,4]✉

High-dimensional linear regression model is the most popular statistical model for high-dimensional data, but it is quite a challenging task to achieve a sparse set of regression coefficients. In this paper, we propose a simple heuristic algorithm to construct sparse high-dimensional linear regression models, which is adapted from the shortest-solution guided decimation algorithm and is referred to as ASSD. This algorithm constructs the support of regression coefficients under the guidance of the shortest least-squares solution of the recursively decimated linear models, and it applies an early-stopping criterion and a second-stage thresholding procedure to refine this support. Our extensive numerical results demonstrate that ASSD outperforms LASSO, adaptive LASSO, vector approximate message passing, and two other representative greedy algorithms in solution accuracy and robustness. ASSD is especially suitable for linear regression problems with highly correlated measurement matrices encountered in real-world applications.

Detecting the relationship between a response and a set of predictors is a common problem encountered in different branches of scientific research. This problem is referred to as regression analysis in statistics. A major focus of regression analysis has been on linear regression models, which search for a linear relationship between the responses and the predictors. Consider the linear regression model of the following form

$$y = X\beta^0 + \varepsilon, \qquad (1)$$

where $y \in \mathbb{R}^n$ is the response vector, $X = (X_1, \ldots, X_p) \in \mathbb{R}^{n \times p}$ is an $n \times p$ measurement matrix with $X_i = (X_{1i}, X_{2i}, \ldots, X_{ni})^\top \in \mathbb{R}^n$ being the $i$-th column, $\beta^0 = (\beta_1^0, \cdots, \beta_p^0)^\top \in \mathbb{R}^p$ is the vector of $p$ true regression coefficients, and $\varepsilon = (\varepsilon_1, \ldots, \varepsilon_n)^\top \in \mathbb{R}^n$ are random errors with $\mathbb{E}(\varepsilon_i) = 0$. The variance of $\varepsilon_i$ is $\mathbb{E}(\varepsilon_i^2) = \sigma^2$ with $\sigma^2$ being the variance of the noise level ($\sigma$ being the typical magnitude of the noise). Let $s_0$ be the number of nonzero entries in $\beta^0$. We focus on the case where $p > n$ and $s_0 < n$, and the goal is to construct a sparse vector $\beta$ which serves as the best approximation to the hidden truth vector $\beta^0$, given $y$ (the measurement results) and $X$ (the measurement matrix) but with $\varepsilon$ (the noise vector) unknown.

Such linear regression models are widely adopted in many practical applications because of their simplicity and interpretability. With the advancement in measurement technologies, high-dimensional data are nowadays accumulating with fast speed in a variety of fields such as genomics, neuroscience, systems biology, economics, and social science. In these high-dimensional data, the number $p$ of predictors is often larger than the number $n$ of samples or measurements ($p > n$), making the solution of the linear regression problem far from being unique. Additional criteria need to be imposed to reduce the degeneracy of solutions and to select the most appropriate linear regression model. One of the most important criteria is sparsity. Motivated by empirical findings in genomics and other fields, we usually assume that the high-dimensional regression models are sparse, in the sense that only a relatively small number of predictors are important for explaining the observed data[1]. Associated with this sparsity criterion are two highly nontrivial issues in high-dimensional linear regression: (1) variables selection, namely to specify the most relevant predictors; and (2) parameters or coefficients estimation, namely to determine the individual contributions of the chosen predictors. Sparse high-dimensional linear regression has also been studied from the angle of compressed sensing[2].

[1]Center for Applied Statistics, School of Statistics, Renmin University of China, Beijing 100872, China. [2]CAS Key Laboratory for Theoretical Physics, Institute of Theoretical Physics, Chinese Academy of Sciences, Beijing 100190, China. [3]School of Physical Sciences, University of Chinese Academy of Sciences, Beijing 100049, China. [4]MinJiang Collaborative Center for Theoretical Physics, MinJiang University, Fuzhou 350108, China. ✉email: sunyifan@ruc.edu.cn; zhouhj@itp.ac.cn

In principle, the regression coefficients can be specified by searching for the solution with the least number of nonzero elements, but this non-convex $l_0$ minimization problem is intractable in practice. Over the years a variety of approaches have been proposed to approximate the optimal $l_0$ solution. The existing approaches can roughly be divided into three categories: relaxation methods, physics-inspired message-passing methods, and greedy methods. The basic idea of the relaxation methods is to replace the non-smooth $l_0$-norm penalty with a smooth approximation. Among them the Least Absolute Shrinkage and Selection Operator (LASSO)[3,4], which uses the $l_1$-norm penalty, is the most popular one. LASSO is a convex optimization problem, which can be solved by methods such as LARS[5–7], coordinate descent[8] and proximal gradient descent[9]. However, due to the over-shrinking of large coefficients, LASSO is known to lead to biased estimates. To remedy this problem, some alternative methods have been proposed, including multi-stages methods such as adaptive LASSO[10] and the three-stage method[11], and non-convex penalties such as the smoothly clipped absolute deviation (SCAD) penalty[12] and the minimax concave penalty (MCP)[13].

An alternative strategy comes from the approximate message-passing (AMP) methods, which are closely related to the Thouless-Anderson-Palmer equation in statistical physics that is capable of dealing with high-dimensional inference problems. They have shown remarkable success in sparse regression and compressed sensing[14–17]. However, the convergence issue limits the practical application of the AMP methods, especially on problems with highly correlated predictors. Recently, several algorithms such as Generalized AMP (GAMP)[18], SwAMP[19], adaptive damping[20], mean removal[20] and direct free-energy minimization[21] were proposed to fix this problem. Especially, the orthogonal or vector AMP (VAMP) algorithm[22,23] offers a robust alternative to the conventional AMP.

Another line of research focuses on greedy methods for $l_0$ minimization such as orthogonal least squares (OLS)[24] and orthogonal matching pursuit (OMP)[25]. The main idea is to select a single variable vector that has the largest magnitude of (rescaled) inner product with the current residual response vector at each iteration step. A sure-independence-screening (SIS) method based on correlation learning was proposed to improve variable selection[26], and an iterative version of this SIS approach (ISIS) could be adopted to enhance the performance of variable selection[27]. Several more recently developed greedy methods proposed to select several variables at a time, including the iterate hard thresholding (IHT) algorithm[28,29], the primal-dual active set (PDAS) methods[30], and the adaptive support detection and root finding (ASDAR) approach[31].

Most of the above-mentioned approximate methods generally assume that the measurement matrix satisfies some regularity conditions such as the irrepresentable condition and the sparse Riesz condition, for mathematical convenience or good algorithmic performance. Roughly speaking, these conditions require that the predictors should be fully uncorrelated or only weakly correlated. But these strict conditions are often not met in real-world applications. As such, it is desirable to develop an efficient and robust method applicable for more general correlation structures of the predictors. Recently, the shortest-solution guided decimation (SSD) algorithm[32] is proposed as a greedy method for solving high-dimensional linear regression. Similar to OLS and OMP, at each iteration step SSD selects a single variable as a candidate predictor. The difference is that this selection is based on the dense least-squares (i.e., shortest Euclidean length) solution of the decimated linear equations. Initial simulation results demonstrated that this SSD algorithm significantly outperforms several of the most popular algorithms ($l_1$-based penalty methods, OLS, OMP, and AMP) when the measurement matrices are highly correlated.

Although the SSD algorithm is highly competitive to other heuristic algorithms both for uncorrelated and correlated measurement matrices, a crucial assumption in its original implementation is that there is no measurement noise ($\boldsymbol{\varepsilon} = \boldsymbol{0}$). As we will demonstrate later, when the measurement noise is no longer negligible, the naive noise-free SSD algorithm fails to extract the sparse solution of linear regression. To overcome this difficulty, here we extend the SSD algorithm and propose the adaptive SSD algorithm (ASSD) to estimate the sparse high-dimensional regression models. Compared with the original SSD, the new ASSD algorithm adopts a much more relaxed termination condition to allow early stop. Furthermore and significantly, we add a second-stage screening to single out the truly important predictors after the first-stage estimation is completed.

We test the performance of ASSD both on synthetic data (predictors and responses are both simulated) and on semi-synthetic data (real predictors but simulated responses, using the gene expression data from cancer samples). In comparison with the representative algorithms LASSO, adaptive LASSO (ALASSO), VAMP, and two greedy methods (ASDAR and SIS-LASSO), our extensive simulation results demonstrate that ASSD outperforms all these competing algorithms in terms of accuracy and robustness of variables selection and coefficients estimation. It appears that ASSD is especially suitable for linear regression problems with highly correlated measurement matrices encountered in real-world applications. On the other hand, ASSD is generally slower than these other algorithms, pointing to a direction of further improvement. It may also be interesting to analyze theoretically the SSD and ASSD algorithms.

## Methods

**The shortest solution as a guidance vector.** We first briefly summarize the key ideas behind the SSD algorithm[32]. Consider the singular value decomposition (SVD) of the measurement matrix $\boldsymbol{X}$: $\boldsymbol{X} = \boldsymbol{U}\boldsymbol{D}\boldsymbol{V}^\top$, where $\boldsymbol{U} = (\boldsymbol{u}_1, \ldots, \boldsymbol{u}_n)$ is an $n \times n$ orthogonal matrix, and $\boldsymbol{V} = (\boldsymbol{v}_1, \ldots, \boldsymbol{v}_p)$ is a $p \times p$ orthogonal matrix, and $\boldsymbol{D}$ is an $n \times p$ diagonal matrix of the singular values $\lambda_1 \geq \lambda_2 \geq \ldots \geq \lambda_n$. Here $\{\boldsymbol{u}_1, \ldots, \boldsymbol{u}_n\}$ and $\{\boldsymbol{v}_1, \ldots, \boldsymbol{v}_p\}$ form a complete set of orthonormal basis vectors for the $n$- and $p$-dimensional real space respectively, so the vectors $\boldsymbol{u}_i = (u_{1i}, \ldots, u_{ni})^\top$ satisfy $\boldsymbol{u}_i^\top \boldsymbol{u}_j = \delta_{ij}$, and vectors $\boldsymbol{v}_i = (v_{1i}, \ldots, v_{pi})^\top$ satisfy $\boldsymbol{v}_i^\top \boldsymbol{v}_j = \delta_{ij}$, where $\delta_{ij}$ is the Kronecker symbol: $\delta_{ij} = 0$ for $i \neq j$ and $\delta_{ij} = 1$ for $i = j$. We can express the true coefficient vector $\boldsymbol{\beta}^0$ as a linear combination of the basis vectors $\boldsymbol{v}_j$:

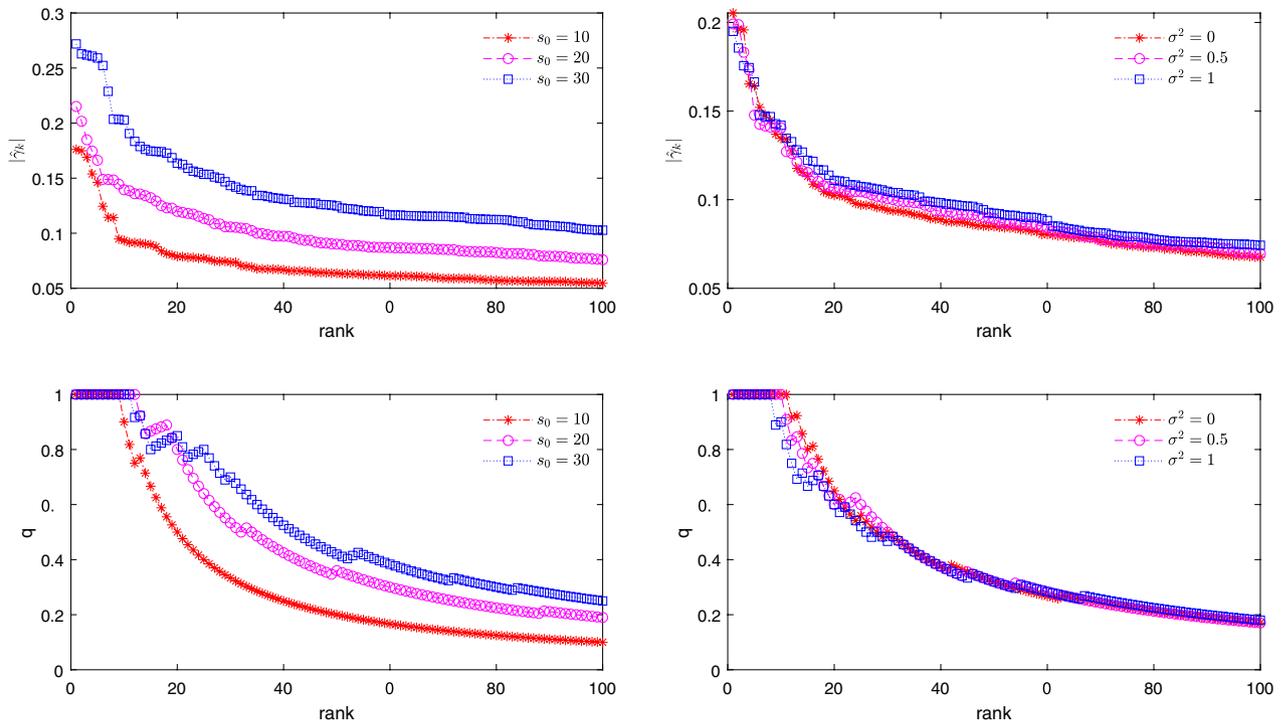

**Figure 1.** Estimated guidance vector $\hat{\gamma}$ on an uncorrelated Gaussian measurement matrix with $p = 1000$ and $n = 200$. Each nonzero coefficient is uniform distributed in [0.5, 1]. Top: rank curves for the estimated guidance vector. Bottom: proportion $q(r)$ of nonzero elements of $\boldsymbol{\beta}^0$ among the $r$ top-ranked indices $i$. In the two subfigures on the left panel, $\sigma^2 = 1$ and $s_0 = 10, 20, 30$; and in the two subfigures on the right panel, $s_0 = 20$ and $\sigma^2 = 0, 0.5, 1$.

$$\boldsymbol{\beta}^0 = \sum_{i=1}^{n} c_i \boldsymbol{v}_i + \sum_{j=n+1}^{p} c_j \boldsymbol{v}_j . \quad (2)$$

Substituting the above expression into the regression function $\mathbb{E}(y|X) = X\boldsymbol{\beta}^0$ of model (Eq. 1), we obtain that

$$\mathbb{E}(y|X) = UDV^\top \boldsymbol{\beta}^0 = \sum_{i=1}^{n} \lambda_i c_i \boldsymbol{u}_i, \quad (3)$$

with the parameter $c_i$ for $i = 1, 2, \ldots, n$ being

$$c_i = \Theta(\lambda_i) \frac{\mathbb{E}(y|X)^\top \boldsymbol{u}_i}{\lambda_i} . \quad (4)$$

Here $\Theta(x)$ is the Heaviside function: $\Theta(x) = 1$ for $x > 0$ and $\Theta(x) = 0$ for $x \leq 0$. We define a vector $\boldsymbol{\gamma}$ as

$$\boldsymbol{\gamma} := \sum_{i=1}^{n} \Theta(\lambda_i) \frac{\mathbb{E}(y|X)^\top \boldsymbol{u}_i}{\lambda_i} \boldsymbol{v}_i. \quad (5)$$

Then

$$\boldsymbol{\beta}^0 = \boldsymbol{\gamma} + \sum_{j=n+1}^{p} c_j \boldsymbol{v}_j. \quad (6)$$

We call $\boldsymbol{\gamma}$ the guidance vector[32]. This vector $\boldsymbol{\gamma}$ is dense and it is not the true coefficient vector $\boldsymbol{\beta}^0$ we are seeking. However, interestingly, this dense vector $\boldsymbol{\gamma}$ does provide information about the locations of nonzero elements of $\boldsymbol{\beta}^0$ (see Fig. 1 and also the earlier empirical observations[32]). To understand this important property of $\boldsymbol{\gamma}$, firstly, we reformulate the matrices $V$ and $D$ as partitioned matrices: $V = (V_1, V_2)$ with $V_1 \in \mathbb{R}^{p \times n}$ and $V_2 \in \mathbb{R}^{p \times (p-n)}$, and $D = (D_1, \mathbf{0})$ with $D_1 = \mathrm{diag}(\lambda_1, \ldots, \lambda_n)$. Then, we have

$$\begin{aligned}\boldsymbol{\gamma} &= V_1 D_1^{-1} U^\top \mathbb{E}(y|X) = V_1 D_1^{-1} U^\top U D V^\top \boldsymbol{\beta}^0 \\ &= V_1 V_1^\top \boldsymbol{\beta}^0.\end{aligned} \quad (7)$$

We define $Q := V_1 V_1^\top$, which is a $p \times p$ symmetric matrix. According to Eq. (7), each element $\gamma_i$ of $\boldsymbol{\gamma}$ is

$$\gamma_i = Q_{ii}\beta_i^0 + \sum_{j\neq i} Q_{ij}\beta_j^0, \tag{8}$$

where $Q_{ii} = \sum_{l=1}^{n} v_{il}^2$, and $Q_{ij} = \sum_{l=1}^{n} v_{il}v_{jl}$. Since $||v_l||_2 = 1$, we may expect that $v_{il} \approx \pm 1/\sqrt{p}$, and thus $Q_{ii} \approx n/p \equiv \alpha$ (here $\alpha$ is the compression ratio). Expecting that $v_{il}$ and $v_{jl}$ are almost independent of each other, we get that $Q_{ij} \approx \pm\sqrt{n}/p$, where $\pm$ means that $Q_{ij}$ is positive or negative with roughly equal probability. Define $\rho := s_0/p$ as the sparsity of $\boldsymbol{\beta}^0$. Because $\boldsymbol{\beta}^0$ is a sparse vector with only $\rho p$ nonzero entries, the summation in the right hand side of Eq. (8) contains at most $\rho p$ terms. Neglecting the possible weak correlations among $Q_{ij}$ ($j \neq i$), we have $\sum_{j\neq i} Q_{ij}\beta_j^0 \approx \pm\frac{\sqrt{n}}{p}\sqrt{\rho p}\,m_0 = \pm\sqrt{\alpha\rho}\,m_0$, where $m_0 = \sqrt{\frac{1}{\rho p}\sum_{i=1}^{p}(\beta_i^0)^2}$ is the mean magnitude of the $\beta_i^0$ coefficients. Putting the above approximations together, we finally get

$$\gamma_i \approx \alpha\beta_i^0 + (\pm\sqrt{\alpha\rho}\,m_0). \tag{9}$$

Notice that the second term in the right hand side of the Eq. (9) is independent of the index $i$. For the element $\gamma_k$ that has the largest magnitude among all the elements of $\boldsymbol{\gamma}$, we expect that the two terms in the right hand side of Eq. (9) have the same sign, and thus it will have a relatively large magnitude. It then follows that the corresponding $\beta_k^0$ is very likely to be nonzero and also $|\beta_k^0| \gtrsim m_0$.

The above analysis offers a qualitative explanation on why the guidance vector $\boldsymbol{\gamma}$ can help us to locate the nonzero elements in the sparse vector $\boldsymbol{\beta}^0$. When there is no noise ($\boldsymbol{\varepsilon} = \mathbf{0}$), this guidance vector is easy to determine and it is the shortest Euclidean-length solution of an underdetermined linear equation. In the presence of measurement noise ($\boldsymbol{\varepsilon} \neq \mathbf{0}$), however, the conditional expectation $\mathbb{E}(\boldsymbol{y}|\boldsymbol{X})$ is unknown. Then we cannot get the exact value of the guidance vector $\boldsymbol{\gamma}$ but can only get an approximate $\boldsymbol{\gamma}$. Consider the estimator

$$\hat{\boldsymbol{\gamma}} = \boldsymbol{V}_1\boldsymbol{D}_1^{-1}\boldsymbol{U}^\top\boldsymbol{y} = \boldsymbol{X}^+\boldsymbol{y}, \tag{10}$$

where $\boldsymbol{X}^+ = \boldsymbol{V}_1\boldsymbol{D}_1^{-1}\boldsymbol{U}^\top$ is the Moore–Penrose inverse of $\boldsymbol{X}$. Notice that $\hat{\boldsymbol{\gamma}}$ is nothing but the shortest length (i.e., minimum $l_2$ norm) least-square solution of linear model (Eq. 1) (hereinafter referred to as the "shortest-solution"). It can be proved that $\hat{\boldsymbol{\gamma}}$ is the best linear unbiased estimator to $\boldsymbol{\gamma}$ (see Supplementary Section A). Combined with the above theoretical analysis, we conjecture that $\hat{\boldsymbol{\gamma}}$ is also helpful for us to guess which elements of the true coefficient vector $\boldsymbol{\beta}^0$ are nonzero. The validity of this conjecture has been confirmed by simulation results. Figure 1 shows the magnitude of elements $\hat{\gamma}_k$ of the estimated guidance vector $\hat{\boldsymbol{\gamma}}$ in descending order (top) and the proportion $q(r)$ of nonzero elements of $\boldsymbol{\beta}^0$ among the $r$ top-ranked indices $i$ (bottom) for datasets generated from models with uncorrelated Gaussian measurement matrices with $n = 200$, $p = 1000$, $s_0 = 10, 20, 30$ and $\sigma^2 = 0, 0.5, 1$. It can be seen that $\hat{\boldsymbol{\gamma}}$ indeed contains important clues about the nonzero elements of $\boldsymbol{\beta}^0$: for the indices $k$ that are ranked in the top in terms of magnitude of $\hat{\gamma}_k$, the corresponding $\beta_k^0$ values have high probabilities to be nonzero. In particular, for the 10 top-ranked indices in the examples of $s_0 = 20$, the corresponding entries in $\boldsymbol{\beta}^0$ are nearly all nonzero.

In practice, the estimated guidance vector $\hat{\boldsymbol{\gamma}}$ can be solved through LQ decomposition or convex optimization which is more efficient than SVD. In our simulation studies, we employ the convex optimization method[32], see Supplementary Section B for the explicit formula.

**Shortest-solution guided Decimation.** Based on the above theoretical analysis and empirical results, we now try to solve the linear model (Eq. 1) through a shortest-solution guided decimation algorithm. Specifically, let $\boldsymbol{\beta} = (\beta_1, \ldots, \beta_p)^\top$ be a $p$-dimensional coefficient vector. Assume that the $k$-th element of the guidance vector $\hat{\boldsymbol{\gamma}}$ has the largest magnitude. If all the other $(p-1)$ elements $\beta_i$ of the vector $\boldsymbol{\beta}$ are known, $\beta_k$ can be uniquely determined as the solution of the minimization problem

$$\begin{aligned}\beta_k &:= \arg\min_{\beta}\left(\boldsymbol{y} - \boldsymbol{X}_k\beta - \sum_{i\neq k}\boldsymbol{X}_i\beta_i\right)^2 \\ &= \frac{\boldsymbol{y}^\top\boldsymbol{X}_k}{\boldsymbol{X}_k^\top\boldsymbol{X}_k} - \sum_{i\neq k}\frac{\boldsymbol{X}_i^\top\boldsymbol{X}_k}{\boldsymbol{X}_k^\top\boldsymbol{X}_k}\beta_i.\end{aligned} \tag{11}$$

Plugging Eq. (11) into model (Eq. 1), we obtain that

$$\boldsymbol{y}' = \boldsymbol{X}'\boldsymbol{\beta}_{-k} + \boldsymbol{\varepsilon}, \tag{12}$$

where $\boldsymbol{\beta}_{-k} = (\beta_1, \ldots, \beta_{k-1}, \beta_{k+1}, \ldots, \beta_p)^\top$, namely the vector formed by deleting $\beta_k$ from $\boldsymbol{\beta}$; $\boldsymbol{X}' = (\boldsymbol{X}_1', \ldots, \boldsymbol{X}_{k-1}', \boldsymbol{X}_{k+1}', \ldots, \boldsymbol{X}_p')$ is an $n \times (p-1)$ decimated measurement matrix with its column vector $\boldsymbol{X}_i'$ being

$$\boldsymbol{X}_i' = \boldsymbol{X}_i - \frac{\boldsymbol{X}_i^\top\boldsymbol{X}_k}{\boldsymbol{X}_k^\top\boldsymbol{X}_k}\boldsymbol{X}_k, \tag{13}$$

and $\boldsymbol{y}'$ is the residual of the original response vector,

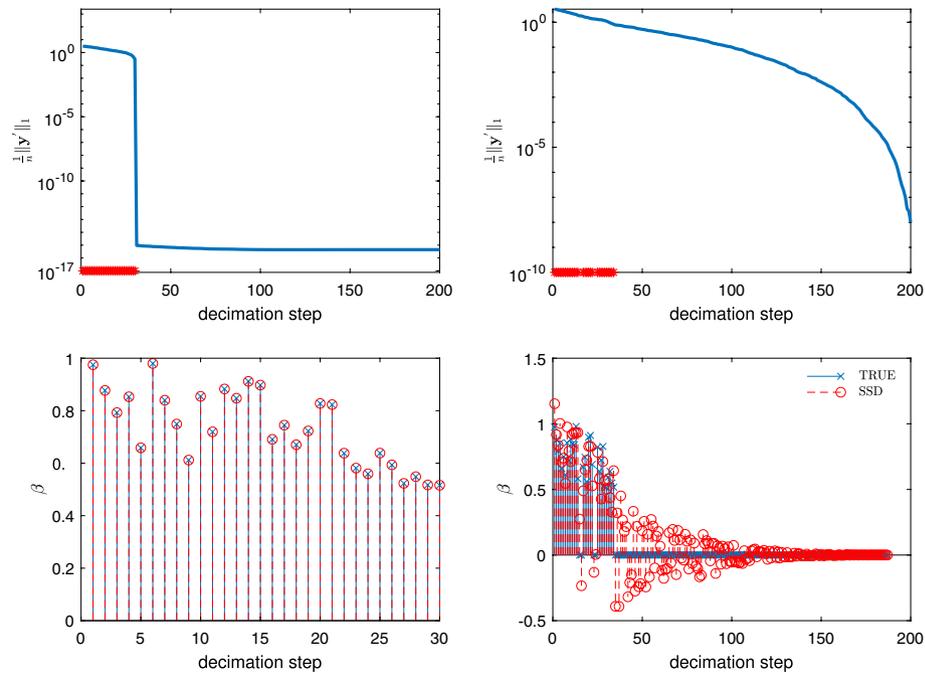

**Figure 2.** Simulation results of SSD with the naive stopping criterion $10^{-5}$ on a single uncorrelated Gaussian measurement matrix ($p = 1000, n = 200$). Top: trace of $l_1$-norm of the residual response vector $\mathbf{y}'$. The red stars on the horizontal axis signify that the identified element $\beta_k^0$ of $\boldsymbol{\beta}^0$, with $k$ being the index of the largest magnitude element $\hat{\gamma}_k$ of $\hat{\boldsymbol{\gamma}}$, is indeed nonzero. Bottom: results of coefficient estimation. The two subfigures on the left panel display the noise-free situation ($\sigma^2 = 0$), and those on the right panel display the noise situation ($\sigma^2 = 1$).

$$\mathbf{y}' = \mathbf{y} - \frac{\mathbf{y}^\top \mathbf{X}_k}{\mathbf{X}_k^\top \mathbf{X}_k} \mathbf{X}_k. \tag{14}$$

Notice that Eq. (12) has the identical form as that of the original linear model (Eq. 1). Therefore, we can obtain the corresponding estimated guidance vector through the least-squares solution of Eq. (12). Then, we repeat the above decimation process (Eqs. 11–14) to further shrink the residual response vector, until the certain stopping criterion is met. Suppose that a total number $L$ of elements of $\boldsymbol{\beta}$ have been picked during this whole decimation process. We can uniquely and easily determine the values of these $L$ elements by setting all the other $(p-L)$ elements to be zero and then backtracking the $L$ constructed equations of the form Eq. (11).

Up to now, this above SSD algorithm is the same as the original algorithm[32]. In the original SSD algorithm, the stopping criterion is that the magnitude of the residual response vector becomes less than a prespecified threshold (e.g., $10^{-5}$). We test the performance of SSD on a single noisy problem instance, to test if this stopping criterion is still appropriate for the noisy situation. Figure 2 shows the trace of the SSD process on two datasets with noise level $\sigma^2 = 0$ (left) and $\sigma^2 = 1$ (right). The two datasets have the identical $200 \times 1000$ measurement matrix $\mathbf{X}$ and the true coefficient vector $\boldsymbol{\beta}^0$ with $s_0 = 30$ nonzero elements, which are sampled from uniform distribution $\mathcal{U}[0.5, 1]$. We see that, for the noise-free situation, the decimation stops (i.e., $\frac{1}{n}\|\mathbf{y}'\|_1 < 10^{-5}$) after $L = 30$ steps (top left) with all the nonzero elements of $\boldsymbol{\beta}^0$ being recovered exactly (bottom left). However, once the noise is added, there is a significant increase in the number of decimation steps ($L = 187$, top right), and the resulting coefficient vector $\boldsymbol{\beta}$ is dense and is dramatically different from $\boldsymbol{\beta}^0$ (bottom right). These results suggest that the stopping criterion used in the original SSD algorithm is no longer appropriate for the linear regression model with noise and needs to be improved.

**Adaptive shortest-solution guided decimation (ASSD).** *Modified stopping criterion.* With an additional examination on the bottom right panel of Fig. 2, we find that during the early steps of the decimation process the identification of the nonzero elements of $\boldsymbol{\beta}^0$ is highly accurate. Specifically, there are only four mistakes in identification in the initial 30 decimation steps. In later decimation steps, however, the index $k$ of the largest-magnitude element $\hat{\gamma}_k$ is no longer reliable, in the sense that the true value of $\beta_k^0$ may be zero. These misidentified elements are too numerous to be corrected by the subsequent backtracking process of SSD, and the resulting coefficient vector $\boldsymbol{\beta}$ is then quite different from $\boldsymbol{\beta}^0$. These observations indicate the necessity of stopping the decimation process earlier.

Firstly, we set an upper bound $L_{\max}$ for the number of decimation steps. It has been established that the true coefficient vector $\boldsymbol{\beta}^0$ cannot be reconstructed consistently with a sample of size $n$ if there are more than $O(n/\ln(n))$ nonzero elements[33]. We therefore take

$$L_{\max} = \frac{n}{\ln n}. \tag{15}$$

We repeat the shortest-solution guided decimation only up to $L_{\max}$ steps. Additionally, we estimate $\boldsymbol{\beta}^0$ by the solution of the $l_1$ minimization problem[34]

$$\min_{\boldsymbol{\beta} \in \mathbb{R}^p} \|\boldsymbol{\beta}\|_1 \quad \text{subject to} \quad \|\boldsymbol{y} - \boldsymbol{X}\boldsymbol{\beta}\|_2 \leq \eta. \tag{16}$$

Under certain conditions on the RIP (restricted isometry property) constant of $\boldsymbol{X}$, the estimation error measured in $l_2$-norm of $\boldsymbol{\beta}^0$ is of the order of $\eta/\sqrt{n}$[34]. Inspired by this insight, we terminate the SSD process earlier than $L_{\max}$ steps once the Euclidean length of the residual response vector (i.e., $\|\boldsymbol{y}'\|_2$) is smaller than a prespecified value $\eta$.

*Second-stage thresholding after SSD.* Even after the early stopping strategy is applied to the decimation process, we find that some of the zero-valued coefficients $\beta_i^0$ are still predicted to be nonzero by the algorithm. To reduce this false-positive fraction as much as possible, we propose a second-stage thresholding procedure to the SSD algorithm. The idea is to manually reset some of the coefficients $\beta_i = 0$ if the value predicted by the SSD algorithm is below a certain threshold value. This refinement procedure turns out to be rather effective in improving the variable selection accuracy.

---

**Algorithm 1** Adaptive Shortest-Solution guided Decimation (ASSD) algorithm for the noisy sparse high-dimensional regression problem (1). This algorithm has one adjustable parameter $\eta$, which is the prespecified precision level.

**Input**: $\boldsymbol{X}' = (\boldsymbol{X}'_1, \ldots, \boldsymbol{X}'_p)$: $n \times p$ residual measurement matrix, initially $\boldsymbol{X}' = \boldsymbol{X}$; $\boldsymbol{y}'$: $n$-dimensional residual response vector, initially $\boldsymbol{y}' = \boldsymbol{y}$; $A$: index set, initially $A = \{1, 2, \ldots, p\}$; $L_{\max}$: maximum number of decimation steps, $L_{\max} = n/\ln n$; $R$: a sufficiently large integer bound, e.g., $R = 20$; $L$: decimation step, initially $L = 0$.

**while** $\|\boldsymbol{y}'\|_2 > \eta$ and $L < L_{\max}$ **do** ▷ decimation with early-stopping

1. Get the least-squares solution $\hat{\boldsymbol{\gamma}} = \{\hat{\gamma}_i : i \in A\}$ for the linear equation $\sum_{i \in A} \hat{\gamma}_i \boldsymbol{X}'_i = \boldsymbol{y}'$.

2. Get leading index $k$ of $\hat{\boldsymbol{\gamma}}$ by criterion $|\hat{\gamma}_k| \geq |\hat{\gamma}_i|$ ($\forall i \in A$), and then delete $k$ from set $A$.

3. Update $L \leftarrow L + 1$, and update $\boldsymbol{X}'_i$ (for every $i \in A$) and $\boldsymbol{y}'$ as

$$\boldsymbol{X}'_i \leftarrow \boldsymbol{X}'_i - \frac{\boldsymbol{X}'^\top_i \boldsymbol{X}'_k}{\boldsymbol{X}'^\top_k \boldsymbol{X}'_k} \boldsymbol{X}'_k, \qquad \boldsymbol{y}' \leftarrow \boldsymbol{y}' - \frac{\boldsymbol{y}'^\top \boldsymbol{X}'_k}{\boldsymbol{X}'^\top_k \boldsymbol{X}'_k} \boldsymbol{X}'_k.$$

**end while**

Set $\beta_i = 0$ for all $i \in A$, and determine the remaining coefficients $\beta_i$ ($i \notin A$) by minimizing $\left\|\boldsymbol{y} - \sum_{i \notin A} \beta_i \boldsymbol{X}_i\right\|_2$.

Sort the $L$ nonzero elements $|\beta_i|$ of $\boldsymbol{\beta}$ in ascending order to compute $\hat{\sigma}$ according to equation (17), and set $\theta_0 = \hat{\sigma}\sqrt{2\ln p}$.

**for** $\tau = 0, 0.01, 0.02, \ldots, R$ **do** ▷ second-stage thresholding

1. Set the actual threshold level $\theta = \tau \theta_0$.

2. If $|\beta_i| < \theta$ ($\forall i \notin A$), then add index $i$ to set $A$.

3. Update the elements of $\boldsymbol{\beta}$ outside set $A$ by solving the minimization problem (19).

4. Get the BIC index according to equation (20) and check if it is the new minimum (if yes, record the vector $\boldsymbol{\beta}$).

**end for**

**Output**: sparse coefficient vector $\boldsymbol{\beta} = (\beta_1, \beta_2, \ldots, \beta_p)^\top$ which has the global minimal value of BIC.

---

Suppose that after early stopping $L$ elements of $\boldsymbol{\beta}$ are assigned with nonzero values, and the indices of all the zero-valued coefficients form a set $A$ (i.e., $\beta_i = 0$ if and only if $i \in A$). We sort the absolute values of these $L$ estimated coefficients in an ascending order (say $|\beta_{r_1}| \leq |\beta_{r_2}| \leq \ldots |\beta_{r_L}|$), and use the first $L/2$ of them to calculate an empirical measure $\hat{\sigma}$ of coefficients uncertainty as

$$\hat{\sigma} = \left( \frac{2}{L} \sum_{j=1}^{L/2} \left( \beta_{r_j} - m \right)^2 \right)^{1/2}, \tag{17}$$

where $m$ means the average value of the considered $L/2$ elements, $m = (2/L) \sum_{j=1}^{L/2} \beta_{r_j}$. Notice $\hat{\sigma}$ is distinct in meaning from the the noise magnitude $\sigma$ of the original model system (Eq. 1). We adopt a data-driven procedure to determine the optimal thresholding level. First we set

$$\theta_0 = \hat{\sigma} \sqrt{2 \ln p} \tag{18}$$

to be the basic thresholding level[35](see also the initial work on thresholding to wavelet coefficients[36]). Next, we take the actual thresholding level $\theta$ to be $\theta = \tau \theta_0$ with $\tau$ taking discrete values. As $\tau$ increases from zero to a relatively large value $R$ (e.g., $R = 20$), the threshold value $\theta$ becomes more and more elevated. At a given value of $\tau$, we first update the index set $A$ by adding some indices $i$ to $A$ if $|\beta_i| < \tau \theta_0$, and then we update the remaining elements of $\boldsymbol{\beta}$ by solving the minimization problem

$$\{\hat{\beta}_i | i \notin A\} = \arg\min_{\{\beta_i | i \notin A\}} \left\| \boldsymbol{y} - \sum_{i \notin A} \beta_i \boldsymbol{X}_i \right\|_2^2. \tag{19}$$

Finally, we compute the BIC (Bayesian Information Criterion) index[33] as

$$\text{BIC} = \frac{1}{2} \left\| \boldsymbol{y} - \sum_{i \notin A} \hat{\beta}_i \boldsymbol{X}_i \right\|_2^2 + p_{\text{nz}} \ln n, \tag{20}$$

where $p_{\text{nz}}$ means the number of nonzero elements in the vector $\boldsymbol{\beta}$, namely $p_{\text{nz}} = p - |A|$. The BIC value is a trade-off between the prediction error and the model complexity. We choose the value of $\tau$ such that BIC achieves the minimum value, and consider the corresponding coefficient vector $\boldsymbol{\beta}$ as the final solution of the linear regression problem (Eq. 1).

*An initial demonstration of ASSD performance.* We summarize the above ideas in the pseudo-code of Algorithm 1. This ASSD algorithm has two parts: decimation with early stopping, followed by refinement by second-stage thresholding.

Let us work on a small example case to better appreciate the working characteristics of ASSD. We generate an $n \times p$ random Gaussian matrix $\boldsymbol{X}$ with $n = 200$ and $p = 1000$, whose elements are i.i.d. $\mathcal{N}(0, 1)$ distributed. The truth coefficient vector $\boldsymbol{\beta}^0$ has $s_0 = 30$ nonzero elements, each of which is sampled from uniform distributions $\mathcal{U}[-1, -0.5]$ and $\mathcal{U}[0.5, 1]$ with equal probability, and 970 zero elements. The response vector $\boldsymbol{y}$ is generated from the linear regression model (Eq. 1) with error level $\sigma^2 = 1$. We compare the performance of ASSD with that of the original SSD which does not conduct early-stopping nor the second-stage thresholding, and that of SSD1, which only adopts early-stopping but skips the second-stage thresholding.

The algorithmic results shown in Fig. 3 reveal that all these three algorithms assigned good approximate values for the nonzero elements of $\boldsymbol{\beta}^0$. SSD has a high false-positive rate (154 of the zero elements of $\boldsymbol{\beta}^0$ are misclassified as nonzero), and early-stopping dramatically reduces this rate (only 8 false-positive predictions in SSD1). By applying the second-stage thresholding, ASSD achieves a zero false-positive rate. In addition, ASSD and SSD1 are more efficient than SSD (SSD, 27.2 s; SSD1, 7.83 s; ASSD, 8.2 s). Overall, the presence of the measurement noise usually renders SSD to produce a solution with a high false-positive rate, and two modifications of ASSD, i.e., a modified early-stopping criterion, coupled with a second-stage thresholding, are proposed to reduce the false-positive rate as much as possible.

## Results

**Model implementation.** To better gauge the performance of ASSD, we compare ASSD with five different methods: LASSO, Adaptive LASSO (ALASSO) , VAMP, SIS+LASSO, and ASDAR. We implemented all these methods in Matlab. Our implementation of LASSO uses the function *lasso*. For ALASSO, we use the LASSO solution $\hat{\boldsymbol{\beta}}^{\text{LASSO}}$ as the initial estimator, and set the weight as $\omega_j = 1/|\hat{\beta}_j^{\text{LASSO}}|$, $j = 1, \ldots, p$. For VAMP, we use the publicly available Matlab package[37]. The algorithm SIS+LASSO first selects $(n \ln n)$ variables based on SIS and then runs LASSO to further reduce the number of falsely identified nonzero coefficients. We implement ASDAR by using the Matlab package *sdar*[31].

For ASSD, we set $R = 20$, $L_{\max} = n/\ln n$ (if not specified) and $\eta = \sqrt{n}\sigma$ (in practical applications, if $\sigma$ (the s.d. of noise) is unknown, we can set $\eta$ to be a small value, e.g. 0.1 as used in Fig. 3). For LASSO, ALASSO, and SIS+LASSO, the tuning parameters are selected by using 10-fold cross validation. For ASDAR, we set $\tau = 5$ and stop the iteration if the number of identified nonzero elements is greater than $L = 0.5n$, or the residual norm is smaller than $\sqrt{n}\sigma$, or the distance of two subsequent solutions (measured in $l_2$-norm) is smaller than 1. For VAMP, a small amount of damping is useful when the measurement matrix is ill-conditioned. We set the dampling parameter to be 0.95. Other parameters of VAMP, including the maximum number of iterations, the tolerance for stopping, are the default values in public-domain GAMPmatlab toolbox[37].

We focus on four metrics for algorithmic comparisons: (1) the relative error (RE) of estimation, defined as $\|\boldsymbol{\beta} - \boldsymbol{\beta}^0\|_2 / \|\boldsymbol{\beta}^0\|_2$; (2) the true positive counts (TP) and (3) the false positive counts (FP) of variable selection, and (4) the CPU time in seconds. In each scenario, we calculate the average and standard deviation of these four metrics over 96 independent runs.

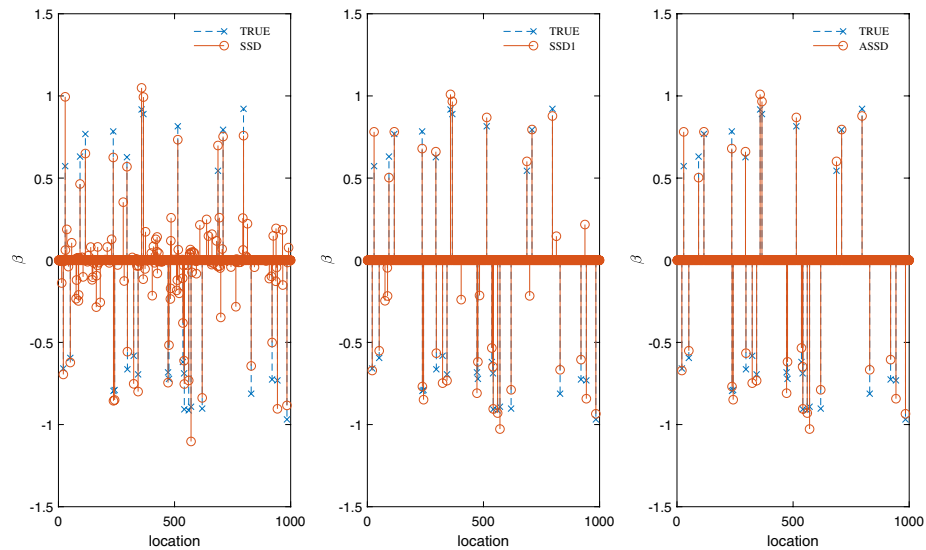

**Figure 3.** Performance comparison on the original SSD (left), SSD1 which is SSD with early-stopping, and ASSD. The measurement matrix is $200 \times 1000$, while the noise level is $\sigma^2 = 1$. The blue crosses mark the 30 nonzero elements of the true coefficient vector $\boldsymbol{\beta}^0$, the red circles mark the elements of the estimated coefficient vector $\boldsymbol{\beta}$. In ASSD, we set $R = 20$ and $\eta = 0.1$.

**Results on three types of measurement matrices.** We first consider different types of measurement matrices $X$ of the same size. In our numerical experiments we set $n = 300$ and $p = 2000$. The number of nonzero coefficients in $\boldsymbol{\beta}^0$ is set to be $s_0 = 40$, with each of them being generated from the uniform distribution $\mathcal{U}[0.5, 1]$. Three types of measurement matrix are considered:

- *Correlated Gaussian matrix*: Each row of the matrix $X$ is drawn independently from $\mathcal{N}(0, \Sigma)$, where $\Sigma_{ij} = \pi^{|j-i|}, 1 \le i, j \le p$ with $\pi = 0$ and $0.7$ corresponding to no and strong correlations.
- *Structured matrix*: The matrix $X$ is the product of an $n \times r$ matrix $X_1$ and an $r \times p$ matrix $X_2$. Both $X_1$ and $X_2$ are random Gaussian matrices whose elements are independently generated from $\mathcal{N}(0, 1)$. The rank $r$ is closely related to the degree of correlation between elements in matrix $X$. When $r \gg n$, the elements in matrix $X$ are weakly correlated or even uncorrelated. As $r$ approaches $n$ from above, the elements in matrix $X$ are more and more correlated. We consider two scenarios: $r = n + 2000 = 2300$ and $r = n + 5 = 305$ corresponding to weakly correlated and highly correlated (or structured) matrices, respectively.
- *Real-world matrix*: We choose the gene expression data from The Cancer Genome Atlas (TCGA) ovarian cancer samples[38] and we use the dataset provided by two earlier studies[39,40]. The dataset is available at https://bioinformatics.mdanderson.org/Supplements/ResidualDisease/. There are 594 samples and 22,277 genes in the original dataset. We randomly subsample the samples and genes to obtain a $300 \times 2000$ measurement matrix $X$.

The response vector $y$ is generated via the linear regression model Eq. (1), in which the random errors are independently generated from normal distribution with means 0 and variance $\sigma^2 = 1$.

*Correlated Gaussian matrix.* Table 1 shows the results on Gaussian matrices. Here and hereinafter, the standard deviations of metrics are shown in the parentheses, and in each column, the numbers in boldface indicate the best performers. It is observed that ASSD has the best performance in variable selection. ASDAR achieves similar performance with ASSD when there is no correlation ($\pi = 0$), but it suffers from identifying more false positives when $\pi$ increases to $\pi = 0.7$. For estimation, ASSD again has the best or close-to-the-best performance compared with VAMP. Although VAMP produces a smaller relative error than ASSD when $\pi = 0$, its performance deteriorates significantly when the correlation $\pi$ is high.

ASSD shows no advantage in speed. ASSD is similar to LASSO and ALASSO in computation time, but it is much slower than ASDAR and VAMP.

*Structured matrices.* Results on the structured matrices are reported in Table 2. We see that when the rank number $r$ of the matrix $X$ is large, i.e. $r = 2300$, VAMP, ASDAR, and ASSD are on the top of the list in metrics for variable selection (TP and FP) and the metric for estimation (RE). As the rank number $r$ approaches $n$ ($r = 305$), ASDAR becomes less accurate in variable selection and coefficient estimation than VAMP and ASSD. The favorable performance of VAMP is not unexpected because it can achieve Bayes-optimal estimation for a class of structured measurement matrix, namely that of the rotationally-invariant matrix. It is encouraging to observe that ASSD performs comparably well in variable selection and estimation, even when the measurement matrix is highly structured.

| $\pi$ | Methods | TP | FP | RE | Time |
|---|---|---|---|---|---|
| 0 | LASSO | **40(0.23)** | 111(22.63) | 3.28E−01 (4.91E−02) | 24.33 (11.06) |
| | ALASSO | **40 (0.32)** | 105 (20.95) | 2.95E−01 (4.09E−02) | 36.41 (18.83) |
| | VAMP | **40 (0)** | 1 (1.74) | **7.31E−02 (8.40E−03)** | **0.12 (0.07)** |
| | SIS+LASSO | 23 (2.16) | 26 (3.04) | 6.89E−01 (5.20E−02) | 0.59 (0.11) |
| | ASDAR | **40 (0)** | **0 (0)** | 7.90E−02 (9.24E−03) | 0.23 (0.10) |
| | ASSD | **40 (0.49)** | **0 (1.34)** | 9.37E−02 (3.95E−02) | 76.82 (52.29) |
| 0.7 | LASSO | **40 (0.20)** | 114 (22.33) | 3.44E−01(5.05E−02) | 27.10 (13.37) |
| | ALASSO | **40 (0.20)** | 104 (20.11) | 3.25E−01 (4.69E−02) | 41.42 (24.76) |
| | VAMP | **40 (0.14)** | 461 (607.04) | 7.76E−00 (5.87E+01) | **0.25 (0.08)** |
| | SIS+LASSO | 18 (2.28) | 21 (3.60) | 7.83E−01 (4.93E−02) | 1.05 (0.11) |
| | ASDAR | **40 (0.71)** | 4 (3.22) | 1.51E−01 (6.13E−02) | 0.37 (0.16) |
| | ASSD | 39 (0.71) | **1 (1.68)** | **1.45E−01 (6.98E−02)** | 72.98 (53.95) |

**Table 1.** Simulation results on Gaussian measurement matrices with $p = 2000$, $n = 300$, $\sigma^2 = 1$, $s_0 = 40$, and $\pi = 0$ or $0.7$. Significant values are in bold.

| $r$ | Methods | TP | FP | RE | Time |
|---|---|---|---|---|---|
| 2300 | LASSO | **40 (0)** | 32 (9.53) | 4.05E−02 (3.59E−03) | 9.40 (2.05) |
| | ALASSO | **40 (0)** | 32 (9.50) | 4.03E−02 (3.17E−03) | 12.66 (2.67) |
| | VAMP | **40 (0)** | **0 (0)** | **1.69E-03 (2.06E-04)** | **0.09 (0.05)** |
| | SIS+LASSO | 22 (2.10) | 26 (3.47) | 7.02E−01 (5.16E−02) | 0.80 (0.05) |
| | ASDAR | **40 (0)** | **0 (0)** | **1.69E−03 (2.06E−04)** | 0.13 (0.03) |
| | ASSD | **40 (0)** | **0 (0)** | **1.69E−03 (2.04E−04)** | 15.76 (2.67) |
| 305 | LASSO | **40 (0)** | 101 (19.64) | 9.41E−02 (2.43E−02) | 30.38 (15.04) |
| | ALASSO | **40 (0)** | 100 (19.47) | 9.31E−02 (2.28E−02) | 44.84 (22.98) |
| | VAMP | **40 (0)** | **0 (0)** | **4.98E−03 (6.40E−04)** | **0.08 (0.05)** |
| | SIS+LASSO | 14 (2.36) | 30 (3.88) | 8.70E−01 (4.43E−02) | 0.99 (0.07) |
| | ASDAR | 38 (5.26) | 30 (41.50) | 1.25E−01 (3.06E−01) | 0.37 (0.26) |
| | ASSD | **40 (0)** | **0 (0)** | 4.99E−03 (6.36E−04) | 17.22 (7.06) |

**Table 2.** Simulation results on structured measurement matrices with $p = 2000$, $n = 300$, $\sigma^2 = 1$, $s_0 = 40$, and $r = 2300, 305$. Significant values are in bold.

| Methods | TP | FP | RE | Time |
|---|---|---|---|---|
| LASSO | 39 (1.84) | 118 (15.87) | 4.80E− 01 (9.02E−02) | 32.10 (3.34) |
| ALASSO | 39 (1.84) | 103 (14.70) | 4.36E−01 (8.12E−02) | 48.07 (5.99) |
| VAMP | **40 (0.32)** | 90 (375.89) | 5.40E−00 (4.11E+01) | **0.18 (0.10)** |
| SIS+LASSO | 7 (2.36) | 20 (4.05) | 9.46E−01 (3.77E−02) | 2.14 (0.73) |
| ASDAR | 35 (6.09) | 18 (16.46) | 4.01E−01 (2.79E−01) | 0.33 (0.16) |
| ASSD | 36 (4.79) | **8 (7.81)** | **3.55E−01 (2.06E−01)** | 31.16 (3.52) |

**Table 3.** Simulation results on a real-world measurement matrix with $p = 2000$, $n = 300$, $\sigma^2 = 1$ and $s_0 = 40$.

*Real-world matrix.* Table 3 shows the results on a real measurement matrix which is a subsample of gene expression data from TCGA ovarian cancer samples. ASSD again has the best performance both in variable selection and in coefficient estimation. LASSO, ALASSO, SIS+LASSO, and VAMP do not work well on the real matrix as they identify too many false positives and produce significantly larger estimation errors. ASDAR is similar to ASSD in terms of estimation error, but it is inferior to ASSD in terms of variable selection. It is indeed quite a remarkable observation that only the ASSD algorithm achieves almost perfect accuracy for this real-world problem instance.

We conduct additional simulation to examine dependence of the proposed ASSD on the distribution of coefficients. First, consider strict sparsity case. The nonzero coefficients are sampled from the uniform distribution $\mathcal{U}[0.5, 1]$ and $\mathcal{U}[-1, -0.5]$ with equal probability; and (b) sampled from uniform distribution $\mathcal{U}[0.2, 1]$. The other settings are the same as above. We use the real-world matrix as a representative and present the results in Supplementary Table S1-S2. Second, consider weak sparsity case, where the $s_0 = 40$ coefficients are sampled from uniform distribution $\mathcal{U}[0.5, 1]$ and other coefficients are set to be 0.001. Supplementary Table S3 reports the results on the real-world matrix. We also consider a larger noise with $\sigma^2 = 1.5$. The nonzero coefficients are sampled from uniform distribution $\mathcal{U}[0.5, 1]$. The simulation results on the real-world matrix are displayed in Supplementary Table S4. Under the above four scenarios, ASSD generally have favorable performance in variable selection and coefficient estimation, in particular, it has the lowest false positives. It is noted that, compared to LASSO and ALASSO, ASSD identifies less true nonzero coefficients. A possible reason is that the adopted stopping criterion is a little bit too strict (i.e., $L$ is too small), and the values of TP are expected to increase with looser stopping criterion. These results demonstrate that ASSD is especially well-suited to scenarios which put a higher value on precision than recall of variable selection.

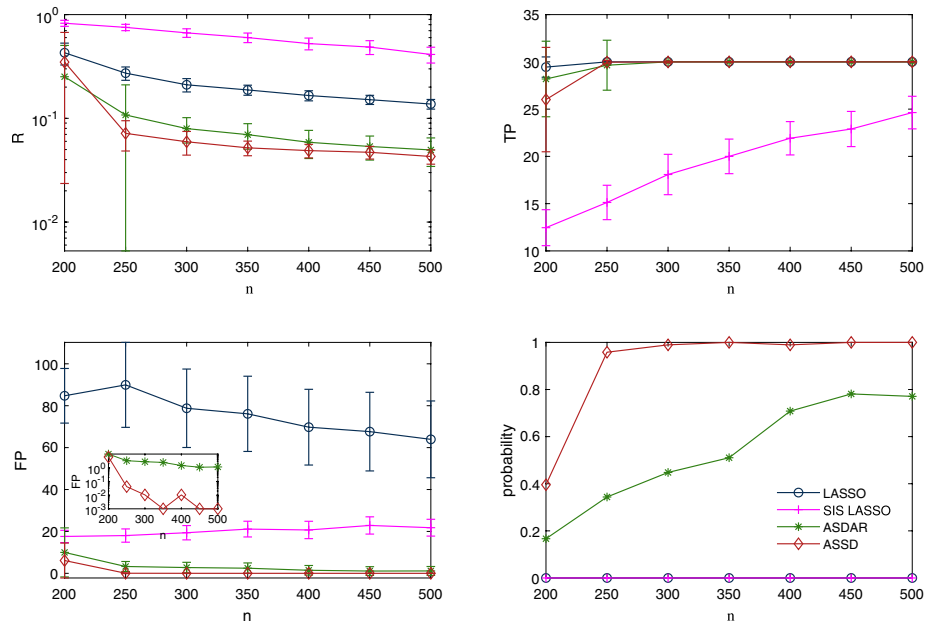

**Figure 4.** Simulation results on a correlated Gaussian measurement matrix ($\pi = 0.7$, $p = 2000$, $s_0 = 30$, $\sigma^2 = 0.5$): influence of the sample size $n$ on relative errors (top left), true positives (top right), false positives (bottom left) with the inset being a semi-logarithmic plot, and probability of exact identification of nonzero coefficients (bottom right).

**Influence of model parameters.** We now investigate more closely the effect of each of the model parameters (the sample size $n$, the number of predictors $p$, and the sparsity level $s_0$) on the performance of LASSO, VAMP, SIS+LASSO, ASDAR, and ASSD. From the above simulation results, we observe that ALASSO performs better than LASSO in variable selection; however, the improvements over LASSO is not large, but with greater computational cost. As such, we do not report the results of ALASSO in this section. The same three types of measurement matrices are examined: Gaussian matrix with $\pi = 0.7$, structured matrix with the rank number $r = n + 5$, and the real-world matrix. The nonzero elements of $\boldsymbol{\beta}^0$ are i.i.d. random values drawn from the uniform distribution over $[0.5, 1]$. We generate the response vector from the linear regression model (1). The random errors are generated independently from $\mathcal{N}(0, 0.5)$. The simulation results are based on 96 independent repeats.

Figure 4 shows the influence of sample size $n$ on the relative errors (top left panel), true positives (top right panel), false positives (bottom left panel), and probability of exact identification of nonzero coefficients (bottom right panel) when the measurement matrix is a correlated Gaussian one. Results obtained on the other two types of measurement matrices can be found as Supplementary Fig. S1 and Supplementary Fig. S2 . (For the real-world matrix and the structured measurement matrices, the results of VAMP are too unstable to be shown here and hereinafter.) As expected, the performances of all the methods improve as $n$ increases. For the real-world and the correlated Gaussian matrices, ASDAR and ASSD perform comparably well in estimation accuracy. However, ASSD performs significantly better than ASDAR in the accuracy of variable selection. Specifically, ASSD can exactly recover the support when $n = 300$, whereas the success probability of ASDAR is only 42%. Similar observations are obtained for the real-world measurement matrix. For the structured measurement matrices, VAMP has the best performance, and ASSD again has close-to-the-best performance compared with ASDAR, LASSO, and SIS+LASSO.

Figure 5 shows the influence of the number of covariates $p$ on the performances of the four methods for a correlated Gaussian measurement matrix. Data are generated from the model with $s_0 = 30$ and $n = 300$. We see that ASSD always produces the lowest relative errors and FP, and the highest TP. In particular, the probability of exactly recovering the support of the true coefficient vector $\boldsymbol{\beta}^0$ of ASSD is higher than that of the other methods as $p$ increases, which indicates that ASSD is more robust to the number of covariates. Similar observations are also made for the other types of measurement matrices as Supplementary Fig. S3 and Supplementary Fig. S4.

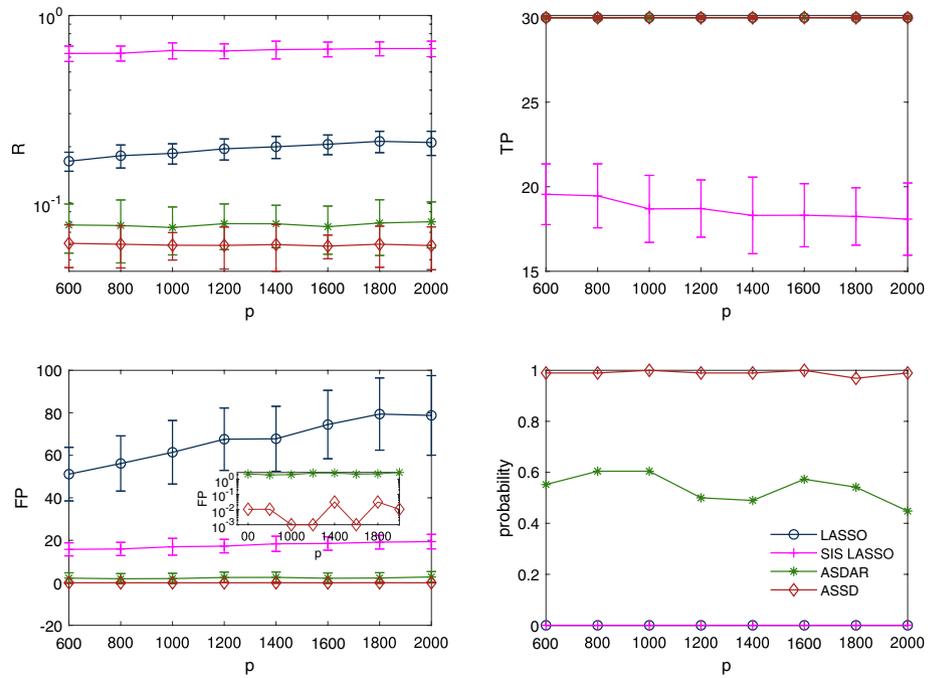

**Figure 5.** Simulation results on a correlated Gaussian measurement matrix ($\pi = 0.7, n = 300, s_0 = 30, \sigma^2 = 0.5$): influence of the covariates number $p$ on relative errors (top left), true positives (top right), false positives (bottom left) with the inset being a semi-logarithmic plot, and probability of exact identification of nonzero coefficients (bottom right).

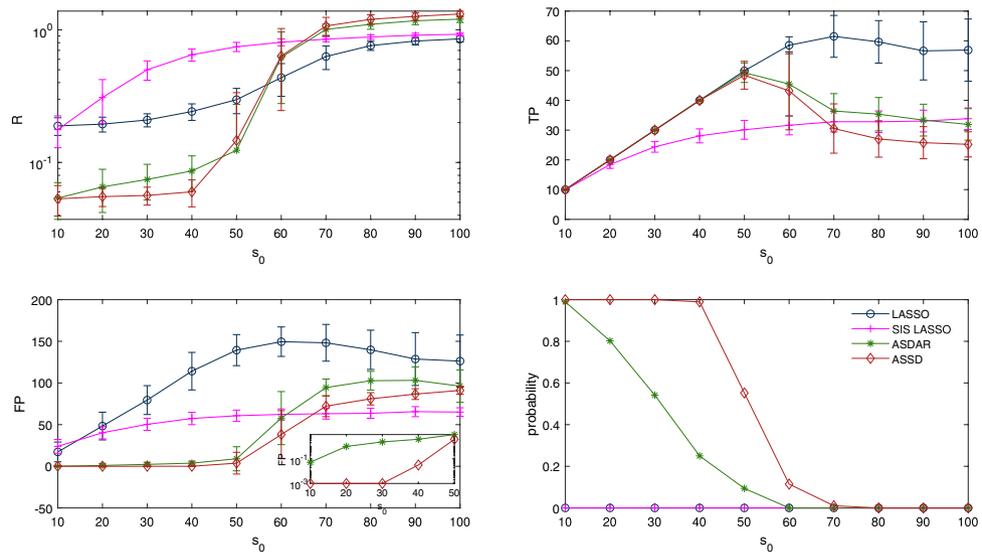

**Figure 6.** Simulation results on a correlated Gaussian measurement matrix ($\pi = 0.7, n = 300, p = 2000, \sigma^2 = 0.5$): influence of the sparsity level $s_0$ on relative errors (top left), true positives (top right), false positives (bottom left) with the inset being a semi-logarithmic plot, and probability of exact identification of nonzero coefficients (bottom right).

The influence of the sparsity level $s_0$ on the performance of the four methods for a correlated Gaussian measurement matrix is presented in Fig. 6. The corresponding results obtained for the other types of matrices are presented as Supplementary Fig. S5 and Supplementary Fig. S6. Data are generated with $n = 300$ and $p = 2000$.

We use $L = 0.5n$ for ASSD and ASDAR as well since the maximum $s_0 = 100$. When the number of nonzero elements $s_0$ increases, the performances of all the four methods become worse. When $s_0$ is small (e.g. $s_0 \leq 40$), ASSD generally has the best performance in the accuracy of estimation and variable selection (i.e., has highest TP and probability of exact identification of nonzero coefficients). However, when $s_0$ is large, ASSD performs worse than some comparison methods. For example, when $s_0 \leq 70$, ASSD produces higher RE and lower TP than LASSO. These results indicate that ASSD is well-suited to "strong sparsity" scenario where the number of nonzero coefficients is small.

In summary, our simulation results demonstrate that the proposed ASSD is more accurate and robust in variable selection and coefficient estimation than LASSO, VAMP, SIS+LASSO, and ASDAR. This ASSD algorithm is a promising heuristic method for highly correlated random and real-world measurement matrices.

## Discussion

In this paper, we proposed the adaptive shortest-solution guided decimation (ASSD) algorithm to estimate high-dimensional sparse linear regression models. Compared to the original SSD algorithm which is developed for linear regression models without noise[32], the ASSD algorithm takes into account the effect of measurement noise and adopts an early-stopping strategy and a second-stage thresholding procedure, resulting in significantly better performance in variables selection (which columns $X_i$ are relevant) and coefficients estimation (what are the corresponding regression values $\beta_i$). Extensive simulation studies demonstrate that ASSD has favorable performance, and outperforms the comparison methods in variable selection, and is competitive with or outperforms VAMP and ASDAR in coefficient estimation. It is robust to the model parameters, and it is especially robust for different types of measurement matrices such as those whose entries are highly correlated. These numerical results suggest that ASSD can serve as an efficient and robust tool for real-world sparse estimation problems.

In terms of speed, ASSD is slower than VAMP and ASDAR and this is an issue to be further improved in the future. To accelerate ASSD, on the one hand, we can select a small fraction of elements in coefficient vector instead of just one of them in each decimation step, and on the other hand, we can adopt a more delicate early-stopping strategy to further reduce the unnecessary decimation steps. In addition, the rigorous theoretical understanding of ASSD needs to be pursued. We have only considered the linear regression model in this paper. It will be interesting to generalize ASSD to other types of models, such as the logistic model and cox model.

## Data availability
The data supporting this study are provided within the paper.

## Code availability
The ASSD code is available as a Matlab code at Github: https://github.com/sugar-xue/ASSD.

### Acknowledgements

We thank the associate editor and two reviewers for careful review and insightful comments, which have led to a significant improvement of this article. This study was partly supported by the National Natural Science Foundation of China Grants No. 12171479, No. 11975295 and No. 12047503, and the Chinese Academy of Sciences Grants No. QYZDJ-SSW-SYS018 and XDPD15.
The computer resources were provided by Public Computing Cloud Platform of Renmin University of China.

### Author contributions
Y.-F.S., H.-J.Z., and X.Y. conceived research; X.Y. performed research; X.Y., Y.-F. S., and H.-J.Z. wrote and reviewed the manuscript.

### Competing interests
The authors declare no competing interests.

### Additional information
**Supplementary Information** The online version contains supplementary material available at https://doi.org/10.1038/s41598-021-03323-7.

**Correspondence** and requests for materials should be addressed to Y.S. or H.-J.Z.

**Reprints and permissions information** is available at www.nature.com/reprints.

**Publisher's note** Springer Nature remains neutral with regard to jurisdictional claims in published maps and institutional affiliations.